\documentclass[aip,graphicx,reprint]{revtex4-1}

\usepackage{graphicx}
\usepackage{dcolumn}
\usepackage{bm}
\usepackage{amsmath,amssymb}
\usepackage[utf8]{inputenc}
\usepackage[T1]{fontenc}
\usepackage{mathptmx}
\usepackage{booktabs}
\usepackage{hyperref}

\begin{document}

\title{Thermodynamics-Informed Input Reparameterization for Neural Prediction of Real-Fluid Thermodynamic Properties in Supercritical Combustion}

\author{Haoze Zhang}
\affiliation{State Key Laboratory of Turbulence and Complex Systems, College of Engineering, Peking University, Beijing 100871, China}

\author{Han Li}
\email{han\_li@pku.edu.cn}
\affiliation{State Key Laboratory of Turbulence and Complex Systems, College of Engineering, Peking University, Beijing 100871, China}
\affiliation{AI for Science Institute (AISI), Beijing 100080, China}

\author{Ke Xiao}
\affiliation{AI for Science Institute (AISI), Beijing 100080, China}

\author{Yangchen Xu}
\affiliation{State Key Laboratory of Turbulence and Complex Systems, College of Engineering, Peking University, Beijing 100871, China}

\author{Runze Mao}
\affiliation{State Key Laboratory of Turbulence and Complex Systems, College of Engineering, Peking University, Beijing 100871, China}

\author{Zhi X. Chen}
\affiliation{State Key Laboratory of Turbulence and Complex Systems, College of Engineering, Peking University, Beijing 100871, China}
\affiliation{AI for Science Institute (AISI), Beijing 100080, China}

\date{\today}

\begin{abstract}

Real-fluid thermodynamic property evaluation is a major computational cost in supercritical combustion simulations. In the enthalpy-based pressure-correction formulation considered here, the cell-local closure evaluates temperature \(T\), density \(\rho\), and the compressibility coefficient \(\psi\) from the solver-available state \((h,p,\mathbf{Y})\) through enthalpy--temperature inversion and repeated real-fluid equation-of-state evaluations. Neural-network surrogates offer fixed-cost inference, but a direct mapping from \((h,p,\mathbf{Y})\) to \((T,\rho,\psi)\) must represent both the caloric relation between enthalpy and temperature and the non-ideal equation-of-state response, resulting in a complex regression problem. This work introduces a thermodynamics-informed input reparameterization strategy, termed target-aligned input reparameterization (TAIR). TAIR replaces the raw enthalpy coordinate of each property network with a target-matched thermodynamic coordinate: the temperature network uses a temperature estimate obtained by inverting a constant-\(c_p\) ideal-gas mixture enthalpy approximation, whereas the density and compressibility networks use an ideal-gas density estimate. These explicit algebraic transformations use only solver-available variables and species constants, guiding the networks to learn real-fluid departures from ideal-gas-based baselines rather than reconstructing the full closure from raw enthalpy. The method is assessed using supercritical methane--oxygen counterflow flame data against a raw-input baseline and target-inconsistent cross-reparameterization controls. TAIR reduces held-out test-set root-mean-squared error (RMSE) by factors of approximately \(1.5\), \(2.0\), and \(7.5\) for \(T\), \(\rho\), and \(\psi\), respectively. For an unseen-strain-rate flame within the augmented thermodynamic envelope, the corresponding factors are \(3.6\), \(14.5\), and \(6.0\). The target-inconsistent controls perform markedly worse, indicating that the gains arise from thermodynamically matched input design rather than generic preprocessing.

\end{abstract}


\maketitle

\section{Introduction}
\label{sec:introduction}

Liquid-propellant rocket engines and related high-pressure propulsion systems commonly operate at pressures above the critical pressure of one or more propellants; consequently, propellant injection, mixing, and combustion are strongly influenced by real-fluid thermodynamics \cite{Yang1995,Mayer2000,Oefelein2006,Yang2000}. Under these conditions, dense-fluid interactions cause pronounced departures from ideal-gas behavior, including strong nonlinearities in density and thermodynamic derivatives near critical or pseudo-critical states \cite{Poling2001,Bellan2000,Banuti2015}. In reacting-flow simulations, these effects are accounted for through the thermodynamic closure. For the enthalpy-based pressure-correction formulation considered here, the local closure evaluates temperature \(T\), density \(\rho\), and the compressibility coefficient \(\psi\) from the solver-available state \((h,p,\mathbf{Y})\), where \(h\) is the specific enthalpy and \(\mathbf{Y}\) is the species mass-fraction vector. Temperature is required for chemical-kinetic and temperature-dependent property evaluations, density enters the conservation equations, and \(\psi\) enters the pressure-correction equation. Accurate and robust evaluation of this closure is therefore essential for supercritical combustion simulations.

The computational expense of this closure arises from repeated real-fluid state evaluations. Given a local state \((h,p,\mathbf{Y})\), temperature is recovered by solving the nonlinear relation \(h=h(T,p,\mathbf{Y})\) typically through Newton-type iteration \cite{Nguyen2025RealFluidFoam}. Each temperature iteration requires an equation-of-state (EoS) evaluation; for a cubic EoS such as Peng--Robinson \cite{Peng1976}, this includes solving the corresponding cubic relation for the molar volume, from which density follows \cite{Foll2019TabulatedEOS}. In the reference closure considered here, \(\psi\) is evaluated by an isenthalpic finite-difference pressure perturbation, which requires an additional enthalpy inversion and EoS evaluation at the perturbed pressure. Because these operations are repeated in every computational cell and at every time step, real-fluid thermodynamic closure can constitute a major computational cost in supercritical combustion simulations \cite{Koukouvinis2022TranscriticalSprays,Wang2018SupercriticalModeling}.

Several approaches have been developed to reduce this cost. Tabulation methods precompute thermodynamic states and replace online EoS evaluations with interpolation \cite{Liu2014LookupAMR,IAPWS2015SBTL}. Their storage requirement and interpolation complexity, however, increase rapidly with the dimensionality and resolution of the tabulated state space. Near critical or pseudo-critical regions, steep real-fluid property gradients additionally require dense local resolution to maintain accuracy \cite{Kawai2015ArbitraryEOS,Ma2017}. Correlated dynamic evaluation reduces repeated property calculations by reusing evaluations within dynamically constructed groups of thermodynamically similar states \cite{Wang2018SupercriticalModeling,Yang2017Tabulation}, but its efficiency and accuracy depend on the quality of the state-space partitioning. These difficulties become increasingly important for multicomponent reacting mixtures, motivating surrogate models that provide rapid local property evaluation without explicitly constructing a high-dimensional table \cite{Wan2025SCFReview}.

Machine-learning models have consequently been explored for accelerating thermodynamic calculations in both non-reacting and reacting real-fluid systems \cite{Roach2023}. In broader thermodynamic applications, neural networks and related regressors have been used to emulate equations of state and iterative flash calculations for complex fluids \cite{Zhu2020MLEOS,Li2019NVTFlash}. For real-fluid flow simulations, Milan et al.~\cite{Milan2021JCP} developed deep-learning surrogates for property evaluation in complex reacting flowfields, demonstrating the feasibility of replacing repeated thermodynamic calculations with neural inference. Sahranavardfard et al.~\cite{Sahranavardfard2024} subsequently implemented a neural real-fluid property model in OpenFOAM for transcritical LOX/GH$_2$ mixing, addressing practical issues including training-domain selection, output transformation, and solver integration. More recently, Cai et al.~\cite{Cai2025} coupled a real-fluid property network with a chemical-ODE network to accelerate both thermodynamic-property evaluation and chemistry integration in supercritical combustion. Collectively, these studies establish neural surrogates as a promising route for accelerating real-fluid reacting-flow calculations.

The available studies have primarily improved surrogate performance through network architecture, training-data coverage, output transformation, loss formulation, and solver coupling. By comparison, the thermodynamic input coordinates themselves have received relatively little systematic investigation, although they determine the regression map presented to the network. In an enthalpy-based formulation, a direct neural closure naturally inherits the solver interface, $(h,p,\mathbf{Y})\mapsto(T,\rho,\psi)$. This input contains sufficient information to determine the target properties and is convenient for CFD coupling. However, the network must represent both the caloric relation between enthalpy and temperature and the non-ideal EoS response within the same learned mapping. This combination may increase the complexity of the regression problem, particularly for density and compressibility in thermodynamically sensitive high-pressure states \cite{Lacaze2012}. It therefore motivates the hypothesis examined in this work: a low-cost thermodynamic transformation of the input coordinates may improve the learnability and accuracy of neural real-fluid closures.

To examine this hypothesis, we introduce a thermodynamics-informed input-coordinate design strategy, termed target-aligned input reparameterization (TAIR), for neural prediction of real-fluid thermodynamic properties. Instead of providing every property network with the same raw enthalpy coordinate, TAIR replaces \(h\) with a target-matched thermodynamic input. The temperature network uses an estimated temperature \(\tilde{T}\) obtained by inverting an ideal-gas mixture enthalpy approximation, whereas the density and compressibility networks use an ideal-gas-based density estimate $\tilde{\rho}=p/(R_{\mathrm{m}}\tilde{T})$. These quantities serve only as input coordinates; the network outputs remain the real-fluid properties \(T\), \(\rho\), and \(\psi\). The transformed coordinates incorporate the leading ideal-gas dependence of the corresponding targets, leaving the networks to represent the remaining real-fluid departures from the ideal-gas baseline rather than reconstructing the complete closure directly from the raw enthalpy coordinate.

The role of input-coordinate design is isolated by comparing TAIR with a raw-input baseline and target-inconsistent cross-reparameterization controls under identical network architectures, training data, and optimization settings. The assessment considers training behavior, held-out test-set accuracy, transfer to a methane--oxygen counterflow flame at an unseen strain rate within the augmented thermodynamic envelope, and closure-level computational cost. The remainder of this paper is organized as follows. Section~\ref{sec:methods} presents the reference real-fluid closure, the TAIR construction, data generation, and neural-network training. Section~\ref{sec:results} reports the model comparisons, and Section~\ref{sec:discussion} discusses the thermodynamic interpretation and implications of the proposed reparameterization, followed by conclusions in Section~\ref{sec:conclusions}.


\section{Methods}
\label{sec:methods}

\subsection{Reference Real-Fluid Thermodynamic Closure}
\label{sec:methodsA}

The present work considers the cell-local thermodynamic closure used in the
enthalpy-based pressure-correction formulation of the DeepFlame
\texttt{dfLowMachFoam} solver \cite{DeepFlame2023}. At each thermodynamic update, the closure receives the local solver state
\begin{equation}
    \mathbf{x}_{h}
    =
    (h,p,\mathbf{Y}),
    \label{eq:solver_state}
\end{equation}
where \(h\) is the mass-specific enthalpy variable selected in the present
solver configuration, \(p\) is the local pressure, and
\(\mathbf{Y}=(Y_1,\ldots,Y_{N_s})\) is the species mass-fraction vector. The
same enthalpy definition and reference convention are used by the flow
solver, the Cantera-based reference thermodynamic evaluator
\cite{Goodwin2023}, and the input transformations introduced in
Sec.~\ref{sec:methodsB}.

The reference real-fluid closure defines the state-to-property mapping
\begin{equation}
    \mathcal{G}:
    (h,p,\mathbf{Y})
    \longmapsto
    (T,\rho,\psi),
    \label{eq:closure_mapping}
\end{equation}
where \(T\) is temperature, \(\rho\) is density, and
\begin{equation}
    \psi
    =
    \left(
    \frac{\partial \rho}{\partial p}
    \right)_{h,\mathbf{Y}}
    \label{eq:psi_definition}
\end{equation}
is the compressibility coefficient used in the pressure-correction
formulation.

For a prescribed state \((h,p,\mathbf{Y})\), temperature is recovered by
solving the nonlinear enthalpy relation
\begin{equation}
    f(T)
    =
    h(T,p,\mathbf{Y})-h
    =
    0,
    \label{eq:enthalpy_root}
\end{equation}
where the real-fluid mixture enthalpy is expressed as
\begin{equation}
    h(T,p,\mathbf{Y})
    =
    h^{\mathrm{ig}}(T,\mathbf{Y})
    +
    h^{\mathrm{dep}}(T,p,\mathbf{Y}).
    \label{eq:enthalpy_decomposition}
\end{equation}
Here, \(h^{\mathrm{ig}}\) is the ideal-gas contribution
obtained from the species thermochemical data, whereas
\(h^{\mathrm{dep}}\) is the real-fluid departure contribution
associated with the equation of state. The enthalpy inversion is performed iteratively. In Newton form, the
temperature update can be written as
\begin{equation}
    T^{k+1}
    =
    T^{k}
    -
    \frac{
        h(T^{k},p,\mathbf{Y})-h
    }{
        c_{p}(T^{k},p,\mathbf{Y})
    },
    \label{eq:newton_temperature}
\end{equation}
where
\begin{equation}
    c_{p}(T,p,\mathbf{Y})
    =
    \left(
    \frac{\partial h}
    {\partial T}
    \right)_{p,\mathbf{Y}}
    \label{eq:cp_definition}
\end{equation}
is the mass-specific real-fluid heat capacity at constant pressure and
composition.

At each trial temperature \(T^{k}\), the real-fluid state at
\((T^{k},p,\mathbf{Y})\) is established using the Peng--Robinson equation
of state. The ideal-gas contributions to \(h\) and
\(c_{p}\) are evaluated from the species thermochemistry at
\(T^{k}\), while their departure contributions are evaluated from the
corresponding Peng--Robinson state. Thus,
\(h\) and \(c_{p}\) are evaluated consistently
from the same trial state and do not require separate EoS root solves.

The Peng--Robinson equation of state is written as \cite{Peng1976}
\begin{equation}
p=
\frac{R_uT}{\bar{V}-b_m}
-
\frac{a_m}
{\bar{V}(\bar{V}+b_m)+b_m(\bar{V}-b_m)},
\label{eq:pr_eos}
\end{equation}
where \(\bar{V}\) is the molar volume and \(a_m\) and \(b_m\) are mixture parameters calculated using the classical one-fluid mixing rules. For each trial temperature, Eq.~\eqref{eq:pr_eos} is solved for the molar
volume. Once the enthalpy inversion has converged, the density is retained
from the final Peng--Robinson state as
\begin{equation}
    \rho
    =
    \frac{\bar{W}}{\bar{V}},
    \qquad
    \bar{W}
    =
    \left(
    \sum_{i=1}^{N_s}
    \frac{Y_i}{W_i}
    \right)^{-1},
    \label{eq:density_from_volume}
\end{equation}
where \(\bar{W}\) is the mixture molecular weight.

In the reference implementation, the compressibility coefficient is
evaluated by a first-order backward finite difference at fixed \(h\) and
\(\mathbf{Y}\). Introducing the perturbed pressure
\begin{equation}
    p'=p(1-\epsilon),
    \label{eq:perturbed_pressure}
\end{equation}
gives
\begin{equation}
    \psi
    \approx
    \frac{
        \rho(h,p,\mathbf{Y})
        -
        \rho(h,p',\mathbf{Y})
    }{
        p-p'
    },
    \label{eq:psi_fd}
\end{equation}
where \(\epsilon>0\) is the relative pressure perturbation. Evaluation of the perturbed state requires a second enthalpy--temperature inversion and real-fluid EoS evaluation at fixed \(h\) and \(\mathbf{Y}\).

\begin{figure}[htbp]
	\centering
	\includegraphics[width=0.48\textwidth]{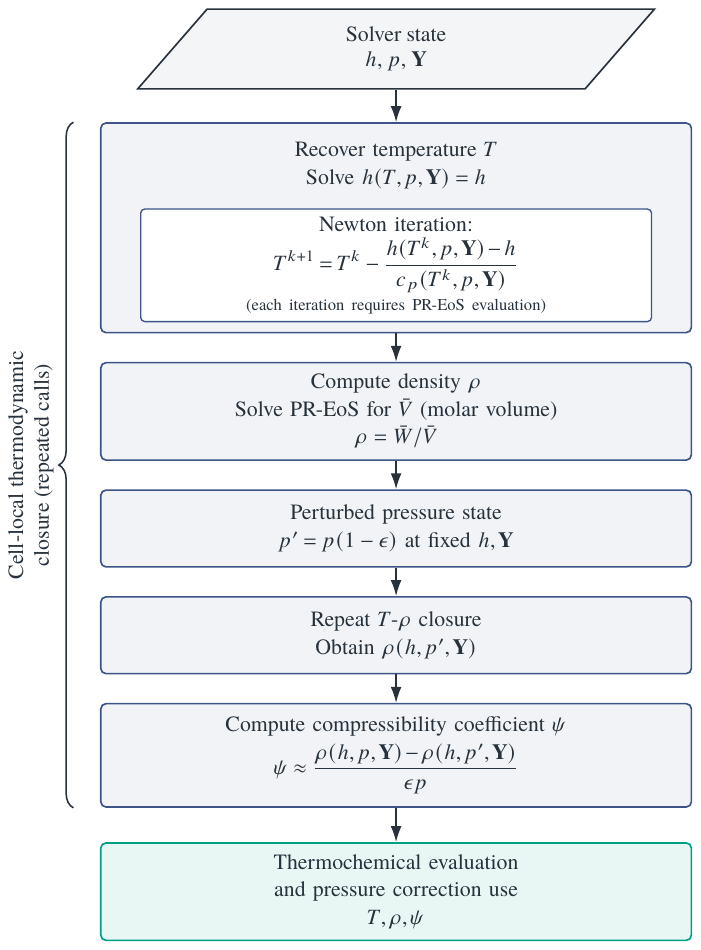}
	\caption{\label{fig:closure_flowchart}
	Schematic of the reference solver-native thermodynamic closure in each computational cell.}
\end{figure}

Fig.~\ref{fig:closure_flowchart} schematically summarizes this cell-local
workflow. Temperature recovery and density evaluation are shown
sequentially for clarity, although they are numerically coupled because
each temperature iteration requires an EoS state evaluation. Evaluation of
\(\psi\) further requires repeating the temperature--density closure at the
perturbed pressure. The reference thermodynamic states underlying
Eqs.~\eqref{eq:enthalpy_root}--\eqref{eq:psi_fd} are evaluated using the
Cantera Peng--Robinson implementation. These calculations provide the supervised targets used throughout this work.

The corresponding raw-input neural surrogate preserves the external
state-to-property interface,
\begin{equation}
    (T,\rho,\psi)
    =
    \mathcal{F}_{\boldsymbol{\theta}}
    (h,p,\mathbf{Y}),
    \label{eq:raw_mapping}
\end{equation}
where \(\mathcal{F}_{\boldsymbol{\theta}}\) denotes the aggregate neural-surrogate mapping. This configuration serves as the raw-input baseline. The TAIR strategy introduced in Sec.~\ref{sec:methodsB} retains \(p\), \(\mathbf{Y}\), and the real-fluid output quantities, but reparameterizes the enthalpy coordinate separately for each target.

\subsection{Target-Aligned Input Reparameterization}
\label{sec:methodsB}

The raw-input baseline defined in Eq.~\eqref{eq:raw_mapping} supplies the
same solver-native enthalpy coordinate \(h\) to all property networks.
Target-aligned input reparameterization (TAIR) modifies only this
thermodynamic coordinate while retaining pressure, composition, network
architecture, and output definitions. The construction follows two
requirements. First, every transformed coordinate must be evaluated
exclusively from the solver-available state \((h,p,\mathbf{Y})\), without using the reference values of \(T\), \(\rho\), or
\(\psi\). Second, the transformation must involve only species-weighted
sums and explicit algebraic operations, so that it does not reintroduce
enthalpy inversion or real-fluid EoS iteration. Its cost therefore scales
linearly with the number of species and is fixed for a prescribed chemical
mechanism.

Because the energy variable used in the present solver configuration is the
mass-specific absolute enthalpy, the ideal-gas mixture enthalpy under the
same thermochemical reference convention can be written as
\begin{equation}
    h^{\mathrm{ig}}(T,\mathbf{Y})
    =
    \sum_{i=1}^{N_s}
    Y_i
    \left[
        h_{f,i}^{\circ}
        +
        \int_{T_0}^{T}
        c_{p,i}^{\mathrm{ig}}(\tau)\,\mathrm{d}\tau
    \right],
    \label{eq:ideal_gas_absolute_enthalpy}
\end{equation}
where \(T_0=298.15\,\mathrm{K}\),
\(h_{f,i}^{\circ}\) is the mass-specific standard enthalpy of formation of
species \(i\) at \(T_0\), and
\(c_{p,i}^{\mathrm{ig}}\) is its mass-specific ideal-gas heat capacity at
constant pressure. All species enthalpies and heat capacities in this
section are expressed on a mass basis before mass-fraction averaging.

A low-cost temperature coordinate is obtained by treating the solver-provided real-fluid enthalpy as an ideal-gas mixture enthalpy and approximating the species heat capacities by their values at \(T_0\):
\begin{equation}
    h^{\mathrm{ig}}(T,\mathbf{Y})
    \approx
    \widetilde{h}^{\mathrm{ig}}(T,\mathbf{Y})
    =
    \sum_{i=1}^{N_s}
    Y_i
    \left[
        h_{f,i}^{\circ}
        +
        c_{p,i}^{\circ}(T-T_0)
    \right],
    \label{eq:constant_cp_enthalpy}
\end{equation}
where $c_{p,i}^{\circ} = c_{p,i}^{\mathrm{ig}}(T_0)$. The thermodynamic coordinate temperature \(\tilde{T}\) is defined by equating the
solver-provided enthalpy to this approximate ideal-gas mixture
enthalpy $ \widetilde{h}^{\mathrm{ig}} (\tilde{T},\mathbf{Y}) = h$.
Solving this equation gives
\begin{equation}
    \tilde{T}
    =
    T_0
    +
    \frac{
        h-\displaystyle\sum_{i=1}^{N_s}Y_i h_{f,i}^{\circ}
    }{
        \displaystyle\sum_{i=1}^{N_s}Y_i c_{p,i}^{\circ}
    }.
    \label{eq:estimated_temperature}
\end{equation}

The quantity \(\tilde{T}\) is a thermodynamic input coordinate rather than a
replacement for the real-fluid temperature. It retains the leading
caloric relation among enthalpy, temperature, and composition,
while omitting the temperature dependence of the species heat capacities
and the real-fluid enthalpy departure. The temperature network therefore
uses
\begin{equation}
    \mathbf{x}_{T}^{\mathrm{TAIR}}
    =
    (\tilde{T},p,\mathbf{Y}).
    \label{eq:tair_temperature_input}
\end{equation}
Pressure is retained because the real-fluid correction to temperature at
fixed enthalpy and composition remains pressure dependent.

For the density-related targets, the thermodynamic coordinate temperature is further combined with the ideal-gas equation of state to obtain the ideal-gas density estimate as
\begin{equation}
    \tilde{\rho}
    =
    \frac{p}
    {R_{\mathrm{m}}\tilde{T}},
    \label{eq:estimated_density}
\end{equation}
where \(R_{\mathrm{m}}\) is the mixture-specific gas
constant.
This coordinate incorporates the leading pressure, temperature, and
mixture-molecular-weight dependence of density, while omitting the
non-ideal EoS contribution.

The use of \(\tilde{\rho}\) for the compressibility network follows from
the same approximate thermodynamic relation. At fixed \(h\) and
\(\mathbf{Y}\), Eq.~\eqref{eq:estimated_temperature} contains no explicit
pressure dependence. Consequently,
\begin{equation}
    \tilde{\psi}
    \equiv
    \left(
        \frac{\partial\tilde{\rho}}
        {\partial p}
    \right)_{h,\mathbf{Y}}
    =
    \frac{1}
    {R_{\mathrm{m}}\tilde{T}}
    =
    \frac{\tilde{\rho}}{p}.
    \label{eq:ideal_compressibility_estimate}
\end{equation}
Thus, because \(p\) is retained as an independent input,
\((\tilde{\rho},p)\) contains the corresponding ideal-gas baseline for the
isenthalpic density response. The density and compressibility networks use
\begin{equation}
    \mathbf{x}_{\rho}^{\mathrm{TAIR}}
    =
    \mathbf{x}_{\psi}^{\mathrm{TAIR}}
    =
    (\tilde{\rho},p,\mathbf{Y}).
    \label{eq:tair_density_inputs}
\end{equation}
Equation~\eqref{eq:ideal_compressibility_estimate} is used only to motivate
the input design; the real-fluid compressibility coefficient \(\psi\)
remains the network output.

The resulting target-specific assignments are summarized in Table~\ref{tab:input_assignments}. The raw-input baseline retains the solver-native enthalpy coordinate for all three targets. TAIR assigns each network the target-matched thermodynamic coordinate that represents the leading ideal-gas dependence of its target. The cross-reparameterization control deliberately exchanges these assignments: \(\tilde{\rho}\) is used for temperature, whereas \(\tilde{T}\) is used for density and compressibility. This control distinguishes the effect of
target-consistent thermodynamic input design from the generic effect of
replacing \(h\) with an analytically transformed variable.

\begin{table}[t]
    \caption{\label{tab:input_assignments}
    First thermodynamic coordinate supplied to each property network.
    Pressure \(p\) and composition \(\mathbf{Y}\) are retained in all
    configurations.}
    \centering
    \begin{tabular}{lccc}
        \hline
        Configuration
        & \(T\) network
        & \(\rho\) network
        & \(\psi\) network \\
        \hline
        Raw-input baseline
        & \(h\)
        & \(h\)
        & \(h\) \\
        TAIR
        & \(\tilde{T}\)
        & \(\tilde{\rho}\)
        & \(\tilde{\rho}\) \\
        Cross-reparameterization
        & \(\tilde{\rho}\)
        & \(\tilde{T}\)
        & \(\tilde{T}\) \\
        \hline
    \end{tabular}
\end{table}

All three configurations predict the absolute real-fluid quantities
\(T\), \(\rho\), and \(\psi\); no residual target is introduced in the
loss function. Therefore, the interpretation that TAIR guides the networks
to learn real-fluid departures from ideal-gas-based baselines refers to
the structure of the input coordinates rather than to an explicit
residual-output formulation. The resulting input--target relations are
examined in Sec.~\ref{sec:results}.

\subsection{Data Generation and Augmentation}
\label{sec:methodsC}

The supervised database consists of input--target pairs
\begin{equation}
    \mathcal{D}
    =
    \left\{
    \mathbf(h,p,\mathbf{Y})^{(n)},
    {(T,\rho,\psi)}^{(n)}
    \right\}_{n=1}^{N}.
    \label{eq:supervised_database}
\end{equation}
It is worth noting that TAIR is applied only after these reference pairs have been generated, as a deterministic preprocessing transformation of the enthalpy coordinate. It therefore does not alter the sampled thermodynamic states, data augmentation, or reference targets. The raw-input, TAIR, and cross-reparameterization models use the same underlying database and differ only in the first thermodynamic coordinate supplied to each property network.

\begin{figure}[t]
	\includegraphics[width=0.48\textwidth]{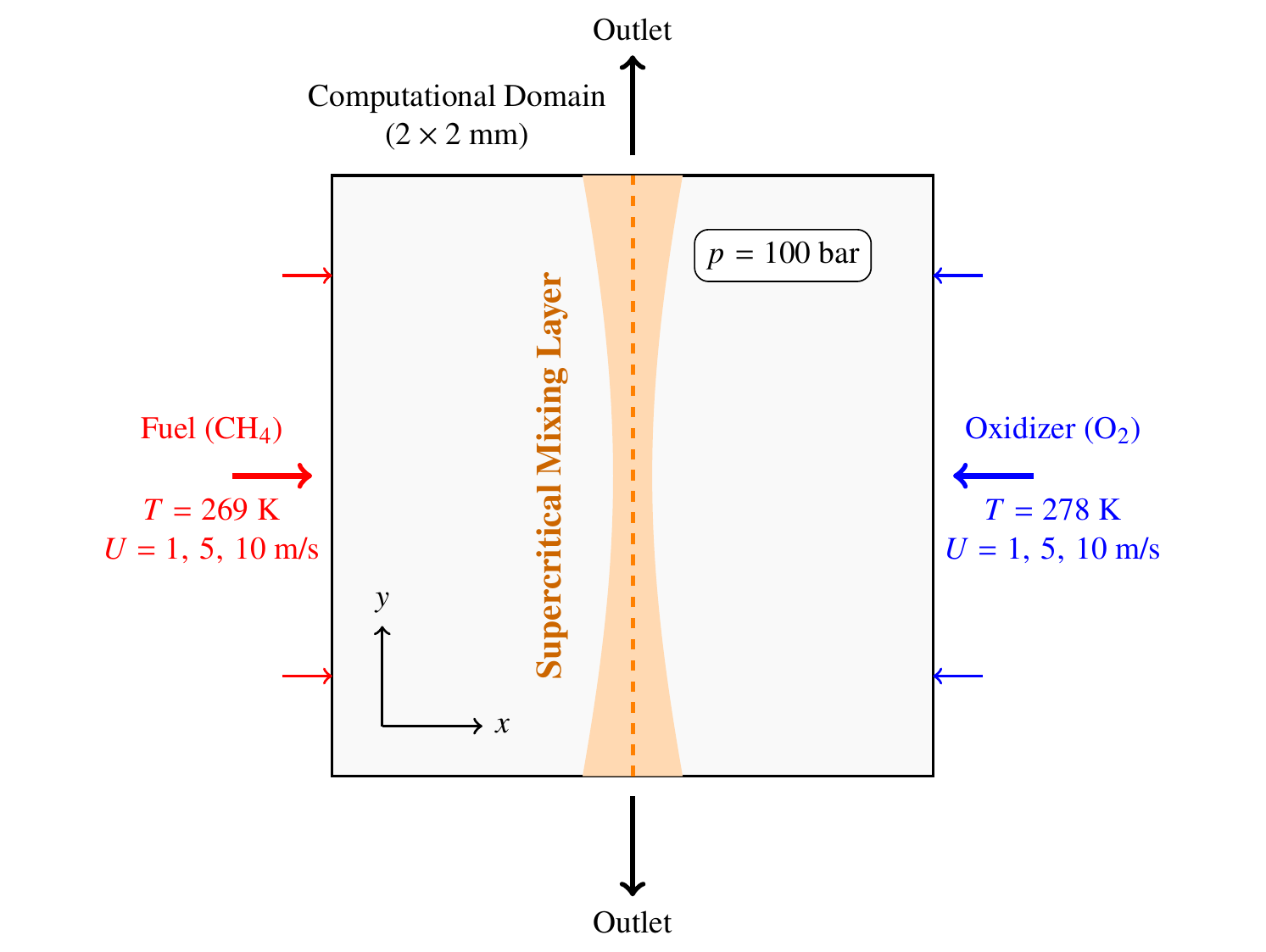}
	\caption{\label{fig:schematic_setup} Schematic of the 2D counterflow diffusion flame configuration used for data generation.}
\end{figure}

The base thermochemical states are sampled from two-dimensional laminar
counterflow diffusion flames \cite{Ribert2008,Monnier2024} computed using the
DeepFlame \texttt{dfLowMachFoam} solver. For each sampled state, the reference thermodynamic properties are evaluated using the Cantera real-fluid closure
described in Sec.~\ref{sec:methodsA}. The counterflow configuration is shown in
Fig.~\ref{fig:schematic_setup}. The computational domain is a
\(2\,\mathrm{mm}\times2\,\mathrm{mm}\) square discretized using a
\(400\times100\) mesh. Pure methane at
\(269\,\mathrm{K}\) and pure oxygen at
\(278\,\mathrm{K}\) are introduced through opposing
boundaries with equal inlet-speed magnitudes. The nominal operating
pressure is \(100\,\mathrm{bar}\), which exceeds the critical
pressures of both pure inlet species.

For equal inlet-speed magnitudes \(U_{\mathrm{in}}\), the nominal strain
rate is defined as $ a = {2U_{\mathrm{in}}}/{L_x}$, where \(L_x=2\,\mathrm{mm}\) is the distance between the opposing inlets. Three flames are simulated at
\(a=1000\), \(5000\), and \(10000\,\mathrm{s^{-1}}\), corresponding
to inlet-speed magnitudes of \(1\), \(5\), and
\(10\,\mathrm{m\,s^{-1}}\), respectively. Chemical kinetics are
represented using the 18-species, 44-reaction reduced mechanism for
high-pressure methane--oxygen combustion developed by Monnier
et al.~\cite{Monnier2022}. For each strain rate, thermochemical states
are sampled from multiple snapshots during the transient evolution
toward a quasi-steady flame, thereby including states from the inlet
streams, mixing layers, reaction zone, and burned-gas region.

\begin{figure}[t]
	\centering
	\includegraphics[width=0.70\columnwidth]{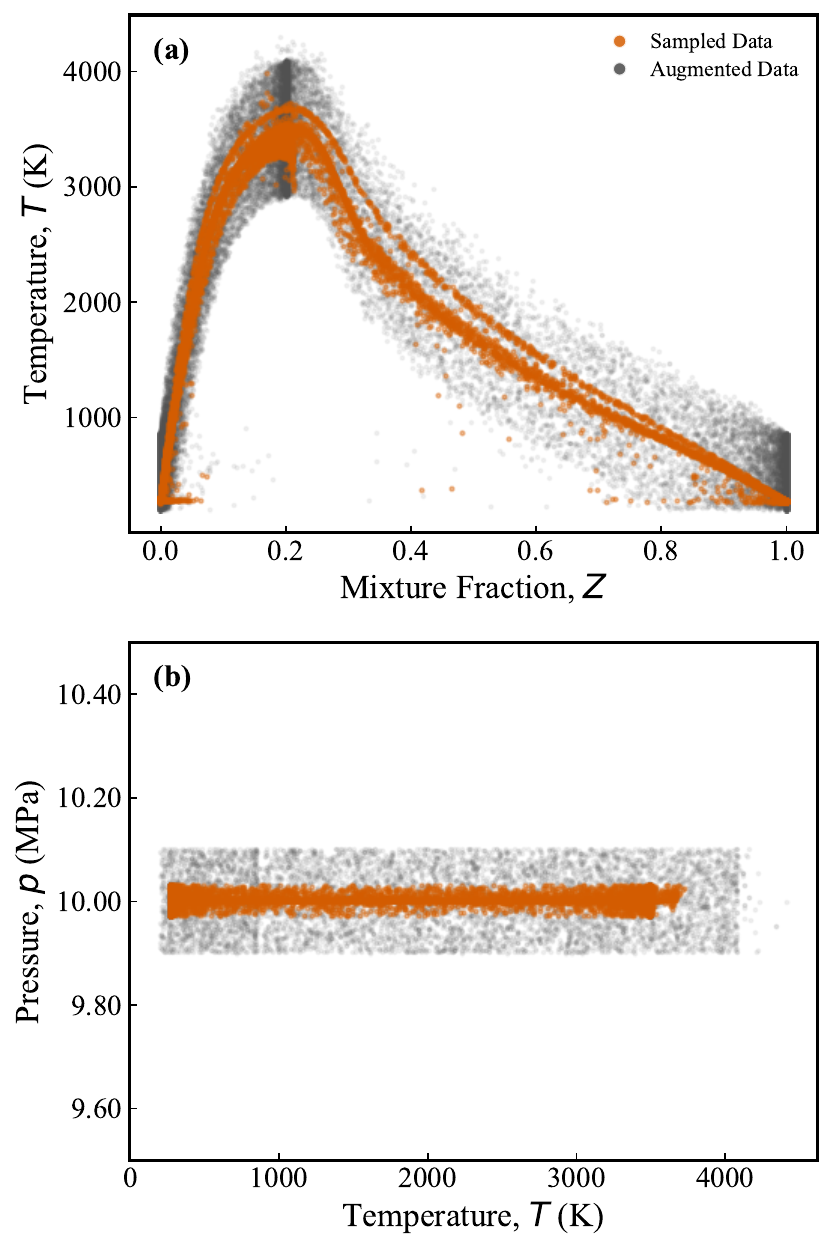}
	\caption{\label{fig:data_coverage} Thermochemical state-space coverage of sampled and augmented flame states. (a) mixture fraction-temperature space and (b) temperature-pressure space.}
\end{figure}

The flame-derived states occupy a relatively narrow portion of the
thermochemical state space. To broaden the thermochemical coverage of the flame-derived database, the physically constrained random-augmentation procedure following our previous work \cite{Li2025} is adopted. The augmentation is performed in \((T,p,\mathbf{Y})\) space rather than by directly perturbing enthalpy: temperature and pressure are randomly varied within prescribed bounds,
while the species mass fractions are perturbed using a common power-law exponent and subsequently renormalized. Candidate states that violate the prescribed temperature, pressure, mixture-fraction, or elemental-composition bounds are discarded. For every accepted augmented state, the mass-specific enthalpy, density, and compressibility coefficient are recomputed using the same
Cantera-based real-fluid closure as that used for the base flame states. Consequently, the augmented input--target pairs remain thermodynamically
consistent. This offline data-generation step is common to all model
configurations and is independent of the subsequent TAIR preprocessing. Fig.~\ref{fig:data_coverage} shows the state-space coverage before
and after augmentation. In the displayed projections, the base flame
states form relatively narrow bands, whereas the augmented samples
populate a broader neighborhood in both the mixture-fraction--temperature and temperature--pressure spaces.

A total of \(1.2\times10^{6}\) accepted samples is retained. The
augmented database is randomly partitioned into
\(1.0\times10^{6}\) training samples,
\(1.0\times10^{5}\) validation samples, and
\(1.0\times10^{5}\) held-out test samples. The resulting test set
therefore evaluates prediction accuracy within the augmented-state
distribution. Transfer across operating conditions is examined
separately using a counterflow flame at the unseen strain rate
\(a=3000\,\mathrm{s^{-1}}\), which is not included in the training,
validation, or test sets but remains within the augmented
thermodynamic envelope.

\begin{figure*}[t]
	\centering
	\includegraphics[width=\textwidth]{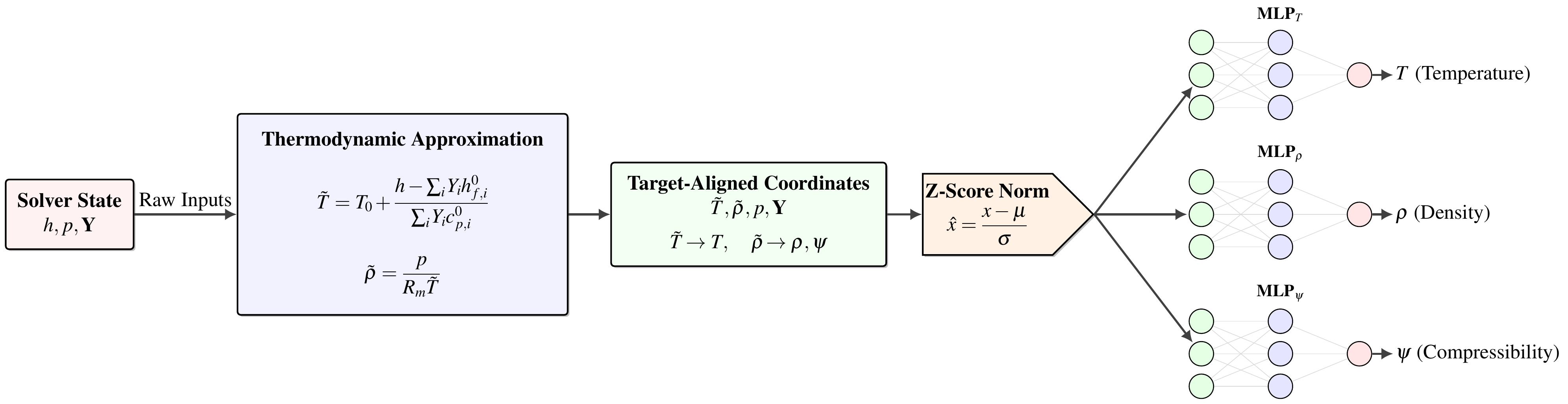}
	\caption{\label{fig:flowchart} Neural-surrogate framework taking the TAIR models as an example.}
\end{figure*}

\subsection{Neural-Network Architecture and Training}
\label{sec:methodsD}

The raw-input, TAIR, and cross-reparameterization models are implemented
using the same neural-network architecture and training procedure. The
three configurations differ only in the thermodynamic coordinate supplied
as the first input to each property network, as summarized in
Table~\ref{tab:input_assignments}. This controlled design isolates the
effect of input reparameterization from those of model capacity, training
data, and optimization.

As illustrated in Fig.~\ref{fig:flowchart}, temperature, density, and
compressibility are predicted using three independent multilayer
perceptrons (MLPs). Each network receives a 20-dimensional input vector
comprising one thermodynamic coordinate, pressure, and 18 species mass
fractions, and returns one scalar real-fluid property. The thermodynamic
coordinate is selected according to the input strategy: the raw-input
model uses \(h\) for all three networks; TAIR uses \(\tilde{T}\) for the
temperature network and \(\tilde{\rho}\) for the density and compressibility networks as presented in Fig.~\ref{fig:flowchart}; and the cross-reparameterization control interchanges these assignments. The target-matched thermodynamic coordinates are therefore used only as network inputs, while the outputs remain the real-fluid quantities \(T\), \(\rho\), and \(\psi\).

Each MLP contains three fully connected hidden layers with 64, 32, and
16 neurons, respectively. Gaussian error linear unit (GELU) activation
functions \cite{Hendrycks2016} are applied after the hidden layers, and
the output layer is linear. The three property networks are trained
independently and do not share parameters. In particular, density and
compressibility are predicted by separate networks rather than deriving
\(\psi\) from the density surrogate, thereby retaining a direct,
fixed-cost evaluation of all three closure quantities.

All input and output variables are standardized using means and standard
deviations calculated from the training set. The same normalization
parameters are subsequently applied to the validation set, held-out test
set, and unseen-strain-rate flame. Each property network is trained using
an \(L_1\) loss in the standardized output space, the Adam optimizer
\cite{Kingma2015}, and a cosine-annealing learning-rate schedule
\cite{Loshchilov2017}. Training is performed with a batch size of 1024
for 1500 epochs. Network predictions are transformed back to physical
units before the accuracy metrics reported in Sec.~\ref{sec:results} are
calculated.

\begin{figure*}[t]
	\includegraphics[width=0.95\textwidth]{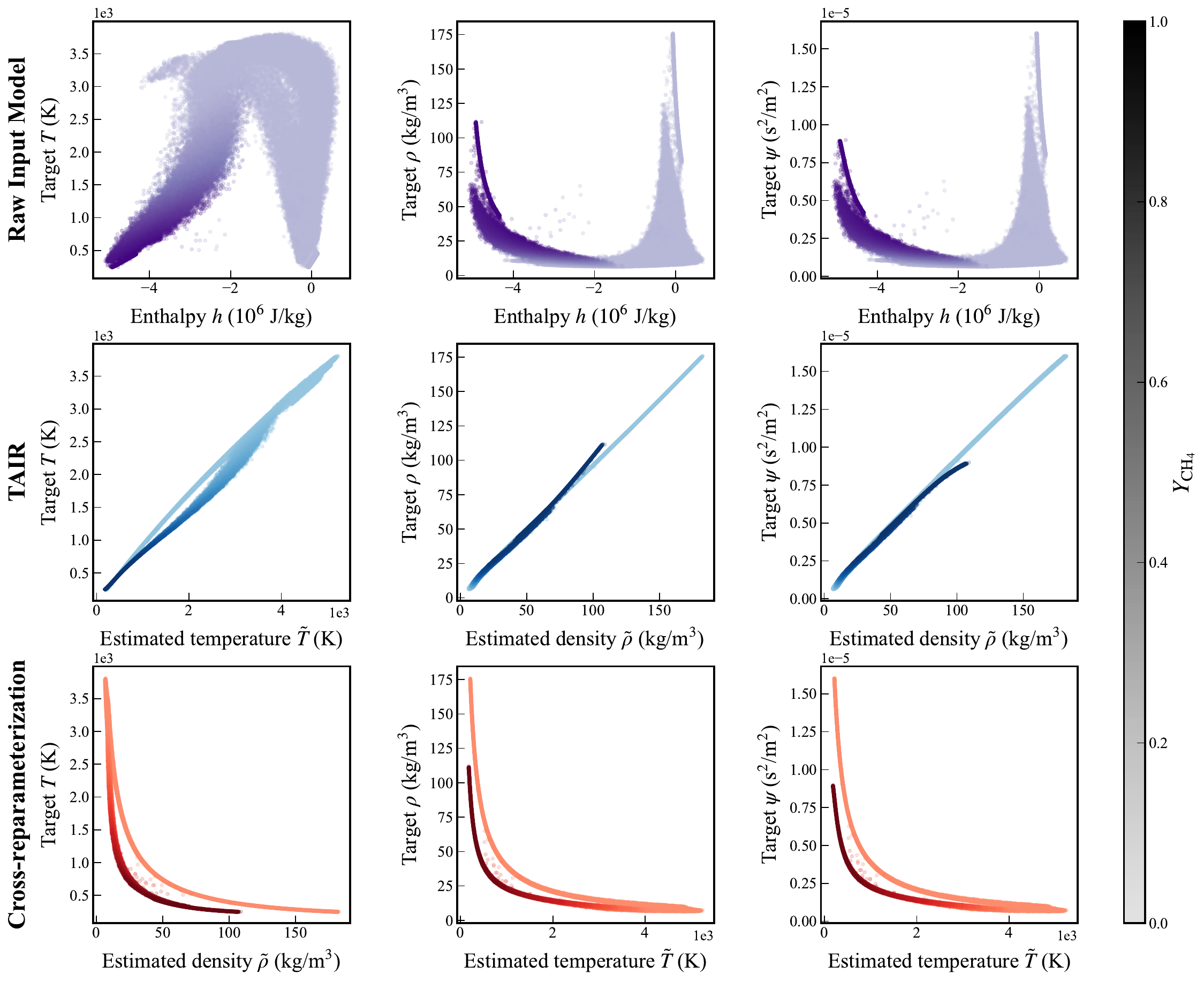}
	\caption{\label{fig:response_surfaces}
	Projected input--target distributions for  temperature, density, and compressibility under the raw-input, TAIR, and cross-reparameterization strategies. Points are colored by the methane mass fraction ($Y_{\mathrm{CH_4}}$).}
\end{figure*}

\section{Results}
\label{sec:results}

The effect of thermodynamic input-coordinate design is examined from five
complementary perspectives: projected input--target relations, training
convergence, held-out test-set accuracy, transfer to an unseen strain-rate
condition, and closure-level computational cost. All comparisons use the
same thermochemical database, property-network architectures, and training
procedure; the only difference among the raw-input, TAIR, and
cross-reparameterization configurations is the thermodynamic coordinate
supplied as the first network input.

\subsection{Projected Input--Target Relations}
\label{sec:projected_relations}

Fig.~\ref{fig:response_surfaces} compares two-dimensional projections of
each target property against the first thermodynamic coordinate used by the
three input configurations. Pressure and the full composition vector remain
network inputs in all cases; consequently, these projections do not
represent the complete high-dimensional regression mapping. They are used
only to examine how the choice of the leading scalar coordinate organizes
the sampled thermodynamic states.

For the raw-input configuration, the projections of \(\rho\), \(T\), and
\(\psi\) against \(h\) are strongly multivalued. States with similar
enthalpy can exhibit substantially different target values because
pressure, composition, and real-fluid departures are not represented by
the enthalpy coordinate alone. The methane-mass-fraction coloring further
shows that much of this projected spread is associated with composition.

With TAIR, the target-matched coordinates produce substantially tighter
and more monotonic projected relations. The mappings
\(\tilde{\rho}\rightarrow\rho\) and
\(\tilde{\rho}\rightarrow\psi\) are close to linear over most of the
sampled range, while the mapping
\(\tilde{T}\rightarrow T\) forms a comparatively narrow monotonic band.
These observations indicate that the target-matched thermodynamic coordinates capture a large part of the leading ideal-gas dependence of their corresponding targets, although residual spread remains because the outputs retain their full
real-fluid dependence.

The cross-reparameterization projections remain organized by analytical
coordinates but exhibit markedly less favorable relations. In particular,
using \(\tilde{\rho}\) for temperature produces an inverse, strongly curved
relation, whereas using \(\tilde{T}\) for density and compressibility
retains broad and nonlinear variations. The comparison therefore suggests
that the benefit is associated with matching the analytical coordinate to
the target thermodynamic dependence, rather than with analytical input
transformation alone.

\subsection{Training Convergence}
\label{sec:resultsA}

\begin{figure*}[t] 
	\includegraphics[width=0.95\textwidth]{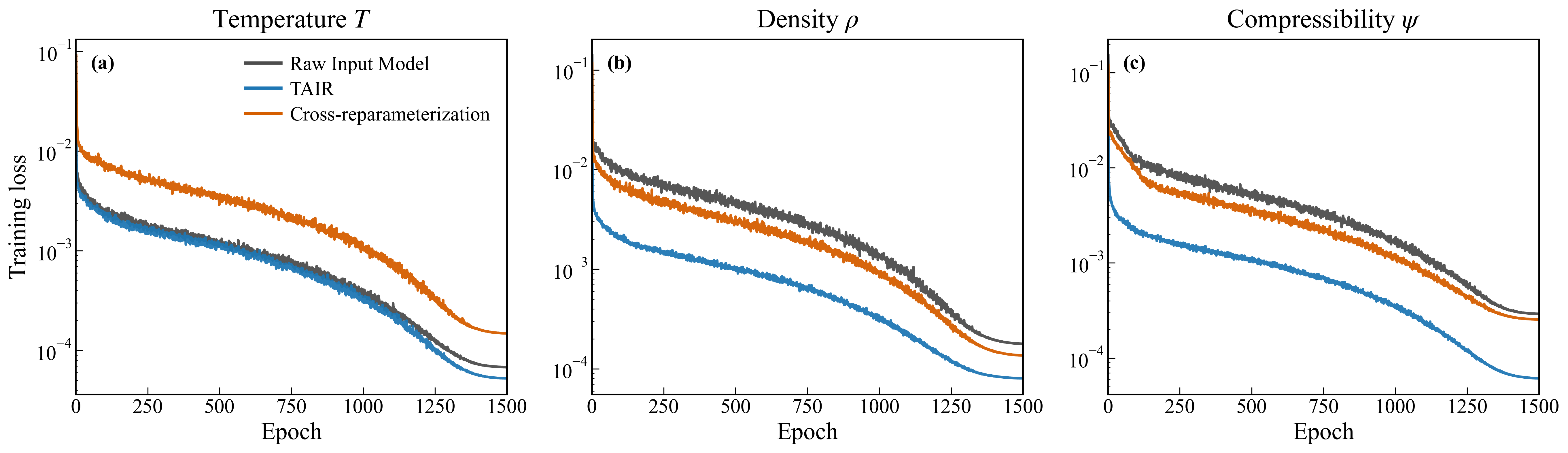}
	\caption{\label{fig:training_loss} Standardized \(L_1\) training-loss histories for the raw-input, TAIR,
    and cross-reparameterization configurations: (a) temperature \(T\),
    (b) density \(\rho\), and (c) compressibility coefficient \(\psi\).}
\end{figure*}

Fig.~\ref{fig:training_loss} compares the standardized \(L_1\) training
losses of the three input configurations. Because each target is
standardized independently, loss magnitudes can be compared among input
configurations for the same property, but not across different
properties.

For density and compressibility, TAIR reaches lower losses from the early
stages of training and maintains this advantage throughout the
optimization. The final density loss is approximately half that obtained
with the raw-input coordinate, while the reduction is more pronounced for
compressibility. For temperature, the raw-input and TAIR models show similar behavior during the initial training stage, but TAIR converges to
the lowest final loss. By contrast, the cross-reparameterization model,
which uses the density-like coordinate \(\tilde{\rho}\) for temperature,
converges more slowly and remains at a substantially higher loss.

The ordering of the converged losses is consistent with the projected
relations in Fig.~\ref{fig:response_surfaces}. Target-matched coordinates
are associated with tighter projected relations and lower attainable
training losses under the same network capacity and optimization
procedure. The cross-reparameterization results further show that replacing
enthalpy by an analytical variable does not by itself guarantee improved
optimization; the relation between the transformed coordinate and the
target property is essential.

\vspace{24pt}

\begin{figure*}[!t]
	\includegraphics[width=0.95\textwidth]{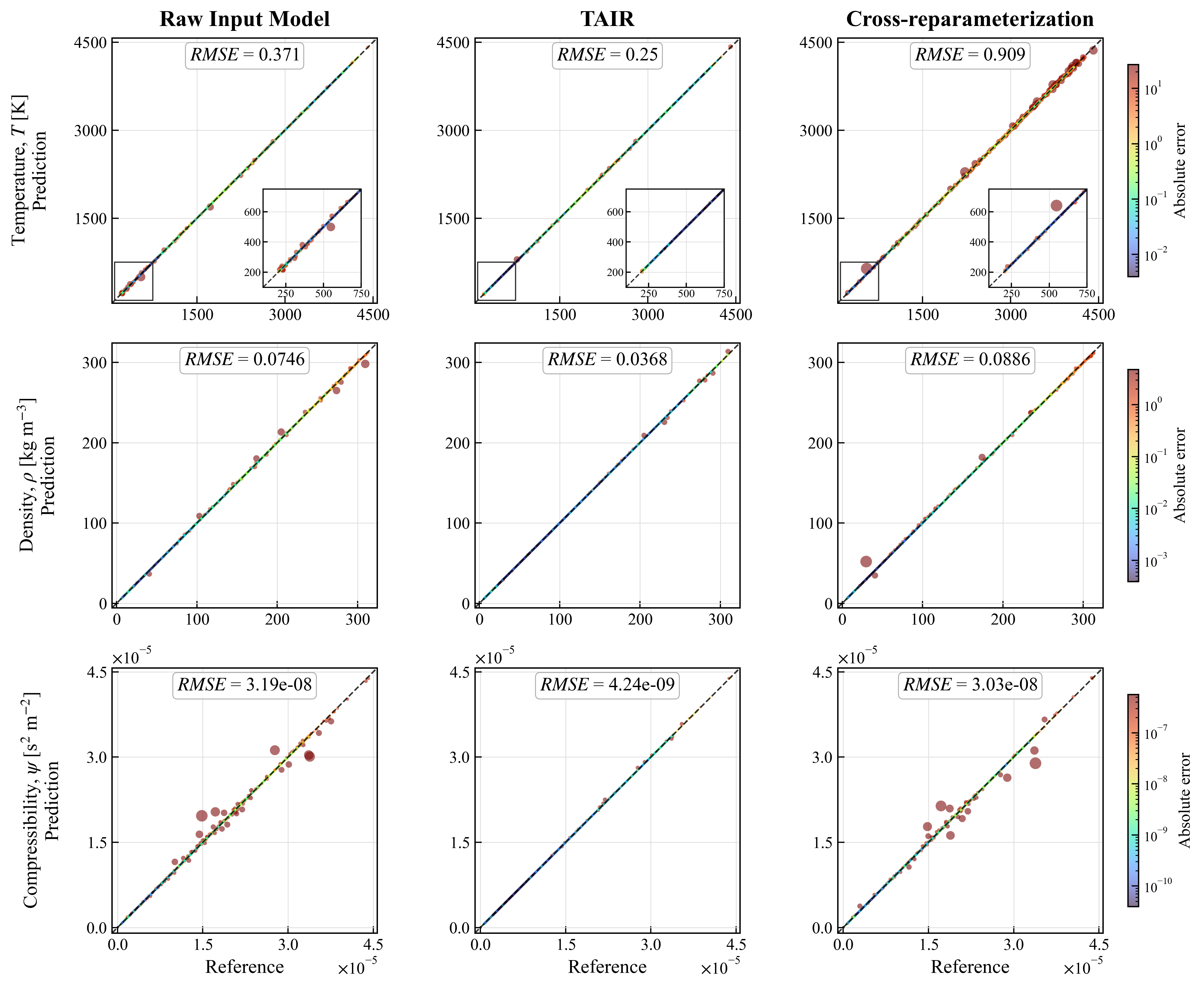}
	\caption{\label{fig:testset_scatter}
    Predicted-versus-reference scatter plots for the held-out test set.
    Rows correspond to the target properties \(T\), \(\rho\), and \(\psi\),
    whereas columns correspond to the raw-input, TAIR, and
    cross-reparameterization models. The dashed lines indicate exact
    agreement. Marker colors and sizes represent absolute errors. Error color scales are normalized
    separately for each target property.
    }
\end{figure*}

\subsection{Held-Out Test-Set Accuracy}
\label{sec:resultsB}

Fig.~\ref{fig:testset_scatter} compares the network predictions with the
reference values on the held-out test subset of the augmented database.
All three configurations produce predictions close to the line of exact
agreement, but their errors differ substantially when quantified by the root-mean-squared error (RMSE) in physical units.

For temperature, the RMSE decreases from \(0.371\,\mathrm{K}\) for the
raw-input model to \(0.250\,\mathrm{K}\) for TAIR, an improvement factor of
approximately \(1.5\). The target-inconsistent
\(\tilde{\rho}\)-to-\(T\) pairing instead increases the RMSE to
\(0.909\,\mathrm{K}\), approximately \(3.6\) times the TAIR value and
\(2.5\) times the raw-input value.

For density, the raw-input model yields an RMSE of
\(0.0746\,\mathrm{kg\,m^{-3}}\), whereas TAIR reduces it to
\(0.0368\,\mathrm{kg\,m^{-3}}\), corresponding to an improvement factor of
approximately \(2.0\). Cross-reparameterization gives
\(0.0886\,\mathrm{kg\,m^{-3}}\), which is less accurate than both TAIR and
the raw-input baseline.

The strongest improvement is obtained for compressibility. TAIR reduces
the RMSE from
\(3.19\times10^{-8}\,\mathrm{s^{2}\,m^{-2}}\) for the raw-input model to
\(4.24\times10^{-9}\,\mathrm{s^{2}\,m^{-2}}\), corresponding to a reduction
factor of approximately \(7.5\). The cross-reparameterization result,
\(3.03\times10^{-8}\,\mathrm{s^{2}\,m^{-2}}\), remains close to the
raw-input error and provides no comparable improvement.

Thus, TAIR gives the lowest held-out error for all three targets. The
cross-reparameterization control does not provide a systematic benefit:
it performs similarly to the raw-input model for compressibility and
degrades the density and temperature predictions. This behavior supports
the interpretation that the improvement originates from target-matched
thermodynamic input design rather than from analytical preprocessing in
general.


\subsection{Transfer to an Unseen Strain Rate}
\label{sec:resultsC}

To assess transfer across operating conditions, an additional counterflow flame is generated at the unseen strain rate \(a=3000\,\mathrm{s^{-1}}\). The assessment is performed \emph{a priori}: the three trained models are queried
on the thermochemical states generated by the reference \texttt{dfLowMachFoam} calculation, without replacing the real-fluid closure during time integration. The comparison therefore evaluates property-prediction transfer to a new flame structure rather than online solver stability or thermodynamic extrapolation.

\begin{figure*}[t]
	\includegraphics[width=0.95\textwidth]{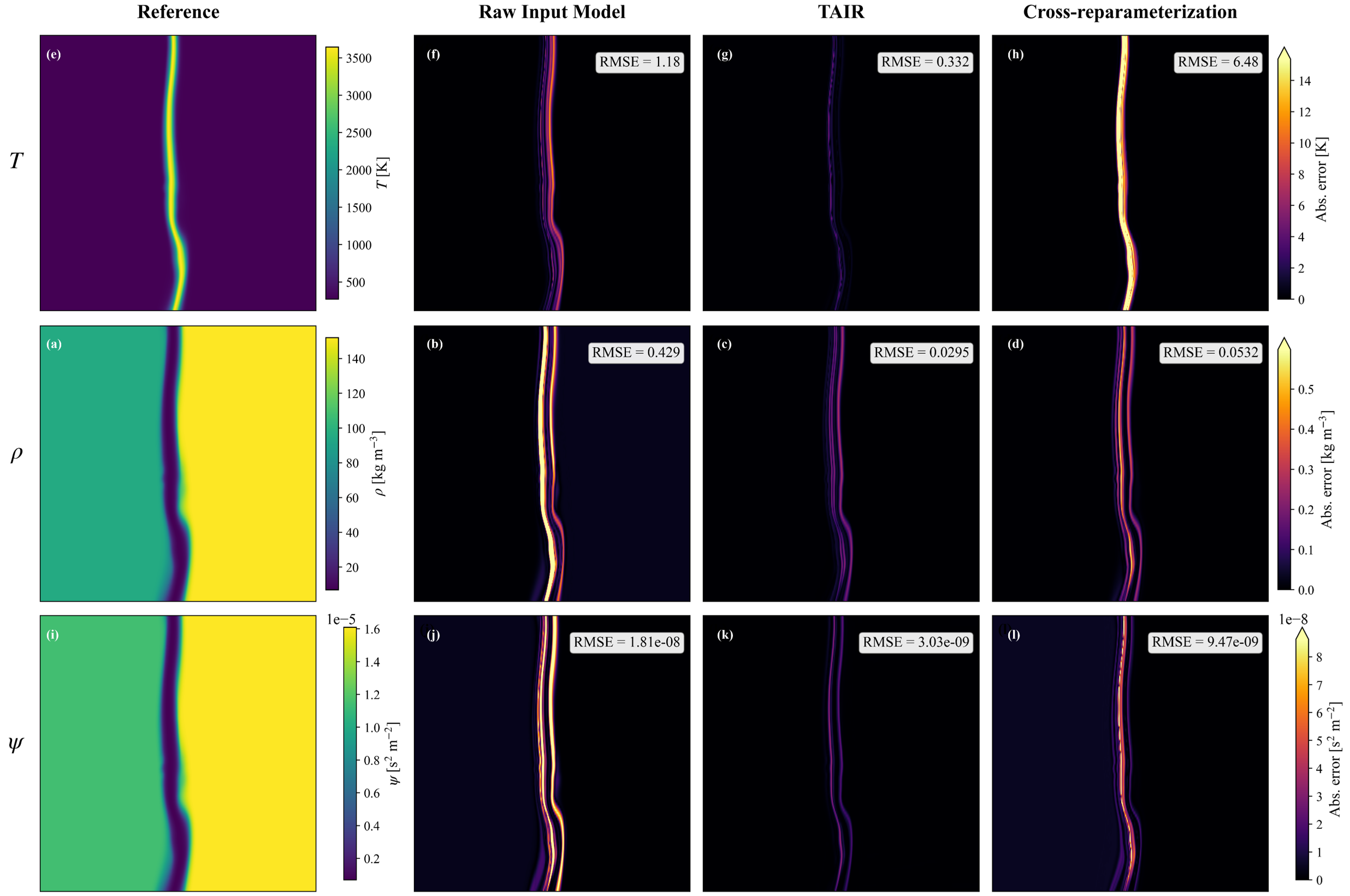}
	\caption{\label{fig:ood_eval} Reference thermodynamic fields and absolute-error contours for the independently generated counterflow flame at the unseen strain rate $a=3000\,\mathrm{s}^{-1}$ and $t=7\times10^{-4}\,\mathrm{s}$.}
\end{figure*}

Fig.~\ref{fig:ood_eval} shows the reference fields and the corresponding
absolute-error maps at
\(t=7\times10^{-4}\,\mathrm{s}\). For temperature, the TAIR RMSE is \(0.332\,\mathrm{K}\), compared with
\(1.18\,\mathrm{K}\) for the raw-input model, giving an improvement factor
of approximately \(3.6\). Cross-reparameterization again performs poorly
for this target, with an RMSE of \(6.48\,\mathrm{K}\). For density, TAIR yields an RMSE of
\(0.0295\,\mathrm{kg\,m^{-3}}\), compared with
\(0.429\,\mathrm{kg\,m^{-3}}\) for the raw-input model and
\(0.0532\,\mathrm{kg\,m^{-3}}\) for cross-reparameterization. The
raw-input-to-TAIR reduction factor is therefore approximately \(14.5\). For compressibility, TAIR reduces the RMSE from
\(1.81\times10^{-8}\,\mathrm{s^{2}\,m^{-2}}\) for the raw-input model to
\(3.03\times10^{-9}\,\mathrm{s^{2}\,m^{-2}}\), corresponding to an
improvement factor of approximately \(6.0\). The
cross-reparameterization RMSE is
\(9.47\times10^{-9}\,\mathrm{s^{2}\,m^{-2}}\), approximately \(3.1\) times
the TAIR value.

For all three properties, the largest errors are confined primarily to
the thin mixing and reaction layer, where the reference thermodynamic
fields exhibit their strongest spatial gradients. Errors in the bulk fuel
and oxidizer streams remain small. The preservation of the TAIR accuracy
advantage at this separately generated flame demonstrates transfer across
strain-rate conditions within the augmented thermodynamic envelope. It
does not, by itself, establish extrapolation beyond that envelope or
stable inline coupling of the surrogate with the pressure-correction
solver.

\subsection{Closure-Level Computational Cost}
\label{sec:resultsD}

The computational cost of the local thermodynamic closure is assessed
using the \(10^{5}\) states in the held-out test subset. The reported
wall-clock time corresponds to completing the evaluation of the full test
set for each method. For every run, timing starts immediately before loading
the test subset and ends after the \(\rho\), \(T\), and \(\psi\) outputs are
materialized.

All timings are obtained using one core of an Intel Xeon Gold 6330 CPU at
\(2.00\,\mathrm{GHz}\). Each method is measured 12 times using a
cyclically balanced execution order. After removing the largest and
smallest values for each method, the reported result is the mean of the
remaining ten measurements.

\begin{table}[t]
\caption{\label{tab:cost_comparison} Thermodynamic closure cost using \(10^{5}\) held-out thermochemical states.}
	\centering
	\begin{tabular}{lcc}
		\toprule
		Method & Mean time (\(\mathrm{s}\)) & Speed-up \\
		\midrule
		Reference (Cantera) & 15.429 & \(1.0\,\times\) \\
		Raw-input NN & 0.159 & \(97.1\,\times\) \\
        TAIR & 0.173 & \(89.4\,\times\) \\
		\bottomrule 
	\end{tabular}
\end{table}

As shown in Table~\ref{tab:cost_comparison}, the reference real-fluid
closure requires \(15.429\,\mathrm{s}\) for the full test set. The raw-input
neural model reduces this cost to \(0.159\,\mathrm{s}\), corresponding to a
speed-up of \(97.1\times\). TAIR requires \(0.173\,\mathrm{s}\) and provides a
speed-up of \(89.4\times\).

The TAIR workflow therefore requires an additional \(0.014\,\mathrm{s}\)
over the full test set relative to the raw-input neural model, corresponding
to an observed overhead of approximately \(8.8\%\). Nevertheless, the
complete TAIR evaluation remains only about \(1.1\%\) of the reference
closure cost. This overhead is therefore small relative to the iterative
thermodynamic calculation that the surrogate replaces.

These measurements quantify closure-level evaluation on the held-out test
set only. They
should not be interpreted as an equivalent speed-up of the complete CFD
simulation, whose wall-clock cost also includes transport equations,
chemical kinetics, pressure correction, communication, and other solver
operations.

\section{Discussion}
\label{sec:discussion}

\subsection{Thermodynamic Interpretation of TAIR}
\label{sec:discussionA}

The target-matched thermodynamic coordinates introduced by TAIR are not approximate substitutes for the real-fluid closure. The estimated temperature \(\tilde{T}\) is constructed by neglecting the real-fluid enthalpy departure and
approximating the ideal-gas species heat capacities as constants at the
reference temperature. The estimated density \(\tilde{\rho}\) further
applies the ideal-gas equation of state to \(\tilde{T}\). These variables
therefore retain the leading caloric or volumetric dependence of the
corresponding targets, while the real-fluid departures remain to be
represented by the neural networks.

TAIR does not introduce additional thermodynamic-state information. At
fixed \(p\) and \(\mathbf{Y}\), the enthalpy \(h\) and
\(\tilde{T}\) are related by an invertible affine transformation. Because
\(p\) and \(\mathbf{Y}\) are retained,
\(\tilde{\rho}=p/(R_{\mathrm{m}}\tilde{T})\) is likewise an invertible
reparameterization over the admissible positive-temperature domain. The
raw-input, TAIR, and cross-reparameterization models therefore contain the
same underlying state information and differ only in how that information
is presented to the finite-capacity networks.

TAIR is thus best interpreted as a deterministic thermodynamic
preconditioning of the input coordinates. The networks continue to predict
the absolute real-fluid quantities \(T\), \(\rho\), and \(\psi\), rather
than explicit residuals. Nevertheless, the target-matched coordinates
expose the leading ideal-gas-based dependence of each target, leaving a
more departure-like input--output relation for neural regression.



\subsection{Target Matching and Local Sensitivity}
\label{sec:discussionB}

The target-dependent improvements are consistent with the structure of the
thermodynamic closure. At fixed \(p\) and \(\mathbf{Y}\), enthalpy is
already directly related to temperature through the caloric equation of
state. The raw enthalpy is therefore a comparatively favorable coordinate
for \(T\), and the transformation to \(\tilde{T}\) mainly accounts for the
dominant formation-enthalpy offset and mixture heat-capacity scaling before
regression. This is consistent with the moderate improvement obtained for
temperature. In contrast, \(\tilde{\rho}\) combines the leading dependence
of density on pressure, caloric state, composition, and mixture molecular
weight. Together with the retained pressure input, it also provides the
ideal-gas baseline for the isenthalpic density response represented by
\(\psi\), as shown in
Eq.~\eqref{eq:ideal_compressibility_estimate}. This is consistent with the
larger improvements obtained for density and compressibility.

\begin{figure*}[t]
    \centering
    \includegraphics[
        width=0.95\textwidth
    ]{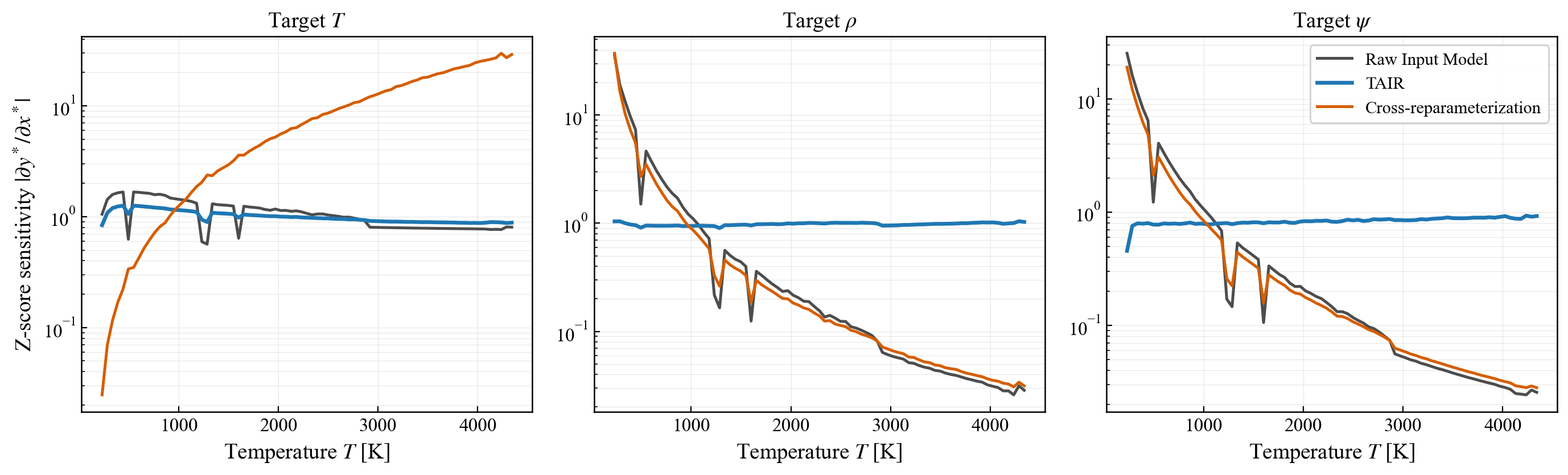}
    \caption{\label{fig:first_coordinate_sensitivity}
    Normalized local first-coordinate sensitivities of the reference
thermodynamic mapping. One representative database state is selected in
each temperature bin as the sample closest to the bin-median temperature.}
\end{figure*}

The cross-reparameterization controls show that the improvement does not
arise from analytical preprocessing alone. Using the density-like
coordinate \(\tilde{\rho}\) for temperature produces a strongly curved
inverse relation, whereas using \(\tilde{T}\) for density and
compressibility omits the leading pressure and mixture-molecular-weight
dependence contained in \(\tilde{\rho}\). The degraded performance of
these target-inconsistent models therefore demonstrates the importance of
matching the transformed coordinate to the thermodynamic dependence of the
target.

Fig.~\ref{fig:response_surfaces} and
\ref{fig:first_coordinate_sensitivity} describe complementary aspects of
this coordinate effect. Fig.~\ref{fig:response_surfaces} provides a
qualitative two-dimensional projection of how each first coordinate
organizes the target states, whereas
Fig.~\ref{fig:first_coordinate_sensitivity} quantifies the corresponding
local slope at representative thermochemical states. For each temperature
bin, one representative database state is selected, and the candidate
first coordinate is locally perturbed while \(p\) and \(\mathbf{Y}\) are
held fixed. The normalized sensitivity is defined as $S_{y,x} = \left|
    {\partial y^{*}}/{\partial x^{*}}
    \right|_{p,\mathbf{Y}}$, where starred quantities denote the standardized variables used during
training. For a monotonic one-dimensional relation, a constant signed
derivative corresponds to an affine mapping. Accordingly, when interpreted
together with the monotonic projections in
Fig.~\ref{fig:response_surfaces}, a narrower variation of \(S_{y,x}\)
indicates a more uniform local input--target scale and a relation that is
closer to affine along the selected coordinate. Such a representation is
generally more favorable for a finite-capacity multilayer perceptron,
consistent with approximation results for smoother target functions and
the spectral preference of standard neural networks for more slowly
varying function components \cite{Yarotsky2017,Rahaman2019,Sokolic2017,pmlr-v32-andoni14}. The more
uniform sensitivities of the target-matched coordinates are therefore
consistent with their tighter projected relations, lower training losses,
and lower prediction errors.

This analysis characterizes only one representative state in
each temperature bin and only the selected first input coordinate. The sensitivity magnitude does not capture derivative-sign changes, cross-coordinate coupling, sample density, or the complete multivariate mapping. Fig.~\ref{fig:first_coordinate_sensitivity} should therefore be interpreted as
a quantitative complement to the projected relations in Fig.~\ref{fig:response_surfaces}, rather than as a complete measure of
mapping nonlinearity or neural-network learnability.

\subsection{Generality and Limitations}
\label{sec:discussionC}

The quantitative results reported in this work are obtained for supercritical methane--oxygen counterflow flames near a nominal pressure of \(100\,\mathrm{bar}\), using the Peng--Robinson equation of state and the thermodynamic envelope represented by the present database. The specific error-reduction factors and computational speed-ups should therefore be interpreted within this validated domain. The unseen-strain-rate case demonstrates transfer to a different flame structure and operating condition within the augmented thermodynamic envelope, rather than unrestricted extrapolation to arbitrary
thermodynamic states.

The input-design principle underlying TAIR, however, is not intrinsically
restricted to the present fuel--oxidizer system or to the three properties
considered here. More generally, when a target quantity admits an
inexpensive analytical or reduced-order approximation based on
solver-available variables, that approximation may be used as a transformed
input coordinate to expose the leading physical dependence of the target
before neural regression. The network is then required to represent the
remaining departure from the physical baseline rather than reconstructing
the complete input--target relation directly from the native solver
variables. This principle may provide a useful design strategy for other
thermodynamic properties, transport quantities, or data-driven closure
terms, provided that an appropriate target-matched baseline can be
identified.

Density and compressibility are predicted using separate networks. This
choice provides direct fixed-cost inference, but does not enforce the
differential consistency condition. Joint training or thermodynamic-consistency regularization may be examined in future work, together with the associated computational trade-offs.

Finally, the flame-field assessment reported here is \emph{a priori}: the
surrogates are queried using states generated by the reference CFD
solution. Stable online coupling with the enthalpy equation and
pressure-correction procedure has not yet been demonstrated. Such a test
must assess error accumulation, pressure--density feedback, conservation,
and excursions outside the training envelope. Extension of the same
input-design principle to transport properties or other closure terms will
similarly require an inexpensive analytical baseline that is physically
matched to the corresponding target.

\section{Conclusions}
\label{sec:conclusions}

This work introduced target-aligned input reparameterization (TAIR) for
neural prediction of real-fluid thermodynamic properties in supercritical
combustion. TAIR replaces the raw enthalpy coordinate of each property
network with a target-matched thermodynamic coordinate: an estimated
temperature derived from a constant-\(c_p\) ideal-gas enthalpy approximation
for temperature prediction, and an ideal-gas density estimate for density
and compressibility prediction. The transformations are explicit,
preserve the solver-available state information, and leave the training
database, network architecture, and real-fluid output definitions
unchanged.

For supercritical methane--oxygen counterflow flames, TAIR reduced the
held-out test-set RMSE by factors of approximately \(1.5\), \(2.0\), and
\(7.5\) for \(T\), \(\rho\), and \(\psi\), respectively, relative to the
raw-input baseline. At an unseen strain rate within the augmented
thermodynamic envelope, the corresponding reduction factors were
approximately \(3.6\), \(14.5\), and \(6.0\). The degraded performance of
the target-inconsistent cross-reparameterization controls shows that the
improvement arises from thermodynamically matched input design rather than
generic analytical preprocessing. TAIR also achieved a closure-level
single-core speed-up of approximately \(89.4\times\) relative to the
iterative Cantera reference. These results demonstrate that low-cost
physical approximations can provide effective input coordinates for
neural real-fluid closures. Future work will assess stable inline coupling
with the pressure-based reacting-flow solver and thermodynamic consistency
between the independently predicted density and compressibility.

\begin{acknowledgments}
	This work was supported by the National Natural Science Foundation of China (Grant Nos. 92270203, 523B2062 and 52276096).
\end{acknowledgments}

\bibliography{aiptemplate}

\end{document}